\pdfoutput=1

\documentclass[12pt]{article}

\usepackage{sbc-template}
\usepackage{graphicx,url}
\usepackage[utf8]{inputenc}
\usepackage[brazil]{babel}

\usepackage{multirow}
\usepackage{mdframed}
\usepackage{booktabs} 
\usepackage{hyperref}
\usepackage{enumitem}

\newcommand{\blue}[1]{\textcolor{black}{#1}}

\title{CTDGSI: A comprehensive exploitation of instance selection methods for automatic text classification\\
\normalsize{VII Concurso de Teses, Dissertações e Trabalhos de Graduação em SI\\
XXI Simpósio Brasileiro de Sistemas de Informação}}

\author{Washington Cunha\inst{1}, Leonardo Rocha\inst{2} (Co-advisor), Marcos Gonçalves\inst{1} (Advisor)}

\address{Department of Computer Science -- Federal University of Minas Gerais -- Brazil
\nextinstitute
Department of Computer Science -- Federal University of São João del Rei -- Brazil
  \email{\{washingtoncunha,mgoncalv\}@dcc.ufmg.br, lcrocha@ufsj.edu.br}
}

\begin{document} 

\maketitle

\begin{abstract}
Progress in Natural Language Processing (NLP) has been dictated by the rule of more: more data, more computing power and more complexity, best exemplified by the Large Language Models. However, training (or fine-tuning) large dense models for specific applications usually requires significant amounts of computing resources. This \textbf{Ph.D. dissertation} focuses on an under-investi\-gated NLP data engineering technique, whose potential is enormous in the current scenario known as Instance Selection (IS). The IS goal is to reduce the training set size by removing noisy or redundant instances while maintaining the effectiveness of the trained models and reducing the training process cost. We provide a comprehensive and scientifically sound comparison of IS methods applied to an essential NLP task -- Automatic Text Classification (ATC), considering several classification solutions and many datasets. Our findings reveal a significant untapped potential for IS solutions. We also propose two novel IS solutions that are noise-oriented and redundancy-aware, specifically designed for large datasets and transformer architectures. Our final solution achieved an average reduction of 41\% in training sets, while maintaining the same levels of effectiveness in all datasets. Importantly, our solutions demonstrated speedup improvements of 1.67x (up to 2.46x), making them scalable for datasets with hundreds of thousands of documents.
\looseness=-1
\end{abstract}

\noindent \textbf{Student Level}: Ph.D., concluded on August 26th, 2024

\noindent \textbf{Board Members}: Franco Maria Nardini (ISTI-CNR); Thierson Couto Rosa (UFG); 	\\Rodrygo Luis Teodoro Santos (DCC-UFMG); 	Anisio Mendes Lacerda (DCC-UFMG)

\noindent \textbf{Dissertation available at}: \url{http://hdl.handle.net/1843/76441}

\noindent \textbf{Publications available at}: \url{http://bit.ly/3WvX1T5}
\section{Introduction}

The exponential growth in data availability has made it challenging to organize and retrieve content effectively. In this context, Automatic Text Classification (\textbf{ATC}) offers a solution by mapping textual documents (e.g., web pages, emails, reviews, tweets, social media messages) into predefined categories of interest. Indeed, ATC models are crucial for emerging applications~\cite{cunha2023tpdr} such as fake news detection, hate speech identification, relevance feedback, sentiment analysis, product review analysis, election vote inference, and assessing public agency satisfaction. As a supervised task, ATC benefits from applications generating large volumes of labeled data, such as social networks like Twitter. Moreover, crowdsourcing~\cite{crowd_2020} and soft labeling~\cite{roy2022soft} further reduce the costs of acquiring labeled datasets.\looseness=-1

Transformer-based architectures, including small and large language models (SLMs and LLMs) such as RoBERTa, GPT-4, and LLama3, represent the state-of-the-art (SOTA) in ATC, achieving outstanding results through pre-training on large datasets and fine-tuning on domain-specific tasks. While zero-shot capabilities are feasible, fine-tuning remains essential for optimal performance, leveraging pre-trained weights for task-specific adaptations~\cite{claudio23}.\looseness=-1

According to Andrew Ng, the success of these (language) models is due to extensive pre-training on massive datasets (e.g., 45TB for GPT-3) and the adaptability of pre-trained models via fine-tuning. This approach enables faster task-specific training compared to starting from scratch~\cite{uppaal2023fine}. However, fine-tuning remains resource-intensive despite being faster than full training, requiring significant computational power and time. For instance, fine-tuning XLNet (one of the datasets considered in our dissertation) on the MEDLINE dataset took 80 GPU hours. Therefore, practical constraints often delay re-training, impacting model performance in scenarios requiring continuous updates, such as fraud detection and recommendation systems.\looseness=-1

Indeed, resource limitations in companies and research groups restrict experimentation with deep learning models, which require extensive computational resources. For example, in this dissertation, we ran 4,000 experiments that took approximately 5,600 hours. Reducing financial, computational, and environmental costs is crucial, as shown by the significant energy consumption and carbon emissions associated with large models. Given increasing data volumes, re-training demands, and environmental concerns, proposing scalable and cost-effective strategies is essential. These include creating efficient deep learning algorithms, using advanced hardware, or improving data preprocessing techniques. This dissertation focuses on the latter, aiming to enhance model performance while reducing costs.\looseness=-1

In~\cite{cunha20}, we proposed exploiting a set of preprocessing techniques in a data transformation pipeline for building cost-effective models. That solution improved effectiveness (e.g., models with higher accuracy) at a much lower cost (e.g., shorter time for ATC model construction). One of the main contributions of this work was the explicit incorporation into the pipeline of an \textbf{Instance Selection} (IS) stage, a promising set of techniques and growing research area that helps to deal with many of the aforementioned issues.\looseness=-1

Specifically, in contrast to traditional Feature Selection approaches~\cite{lu2017feature}, in which the main objective is to select the most informative terms (words), \textbf{Instance Selection} methods are focused on selecting the most representative instances (documents) for the training set~\cite{taxonomy}. The intuition behind this type of algorithm is to remove potentially noisy or redundant instances from the original training set and improve performance in terms of total time training time while keeping or even improving effectiveness. More specifically, IS methods have three main goals: (i) to reduce the number of instances by selecting the most representative ones; (ii) to maintain (or even improve) effectiveness by removing noise and redundancy; and (iii) to reduce the total time for applying an end-to-end model (which includes from traditional preprocessing steps to the model training step). By selecting the most representative instances, IS methods can also potentially remove noise from erroneous annotations. According to these objectives, IS methods must respect three fundamental constraints consisting of \textit{reducing the amount of training without loss of effectiveness and with efficiency gains}. IS methods seek to optimize these three constraints simultaneously.\looseness=-1 

Despite their potential, IS approaches have been understudied in ATC, particularly in deep learning~\cite{barigou2018impact,10.5555/2747013.2747139}. Traditional IS methods developed for other domains include \textbf{CNN}~\cite{hart1968condensed}, \textbf{ENN}~\cite{wilson1972asymptotic}, and \textbf{Drop3}~\cite{wilson2000reduction}. Recent methods, such as \textbf{LSSm}, \textbf{LSBo}~\cite{lssm_lsco_lsbo}, and \textbf{PSDSP}~\cite{psdsp}, have mainly been tested on small tabular datasets with weak classifiers such as KNN. In contrast, text classification datasets are unstructured, larger, and more complex, featuring high dimensionality and skewness. Given the high computational costs of deep learning transformer models with large training datasets, presenting an ideal scenario for applying IS techniques.\looseness=-1

\section{Hypothesis and Research Questions}

The main hypothesis (H1) of this Ph.D. dissertation is:
\vspace{0.1cm}
\begin{mdframed}
\centering
\textbf{H1: It is possible to simultaneously reduce data, maintain model quality, and improve time for fine-tuning ATC models through IS methods.}
\end{mdframed}

In order to confirm this hypothesis, we propose \textbf{three} research questions for our Ph.D. dissertation. In sum, \textbf{RQ1} aims at evaluating the traditional IS approaches shall be present in Section~\ref{sec:review} in the context of the ATC task concerning the posed IS methods tripod constraints: reduction, effectiveness, and efficiency. Next, considering the literature gap, \textbf{RQ2} aims at demonstrating the feasibility of proposing a novel IS framework and by showing how it can accommodate different requirements posed by distinct scenarios, mainly those associated with big data. Finally, \textbf{RQ3} aims to investigate the capability of each IS method to handle and effectively remove noisy instances. The answer to this question also motivated us to demonstrate the feasibility of proposing a novel extended IS framework capable of removing simultaneously redundant and noisy instances from the training set. Next, we present each of the RQs considered in this Ph.D. dissertation.\looseness=-1

\begin{description}
\item[\emph{RQ1.}] \textbf{\textit{What is the impact of applying traditional IS methods in the ATC context regarding the posed constraints?}} RQ1 aims to evaluate the traditional IS approaches in the context of the ATC task, focusing on the tripod constraints of the posed IS methods: reduction, effectiveness, and efficiency. \looseness=-1
\item[\emph{RQ2.}] \textbf{\textit{Can a novel instance selection method focused on redundancy removal overcome the limitations of existing IS  methods to achieve the tripod restrictions in the ATC scenario?}} This RQ aims to demonstrate the feasibility of proposing a novel IS framework and by showing how it can accommodate different requirements posed by distinct scenarios, mainly those associated with big data.\looseness=-1
\item[\emph{RQ3.}] \textbf{\textit{Is it possible to extend the previous proposal to not only remove redundancy but also remove noise, enhancing the level of quality considering all tripod criteria? }} The objective of this RQ is to demonstrate the feasibility of proposing a novel extended IS framework capable of removing simultaneously redundant and noisy instances from the training set.
\looseness=-1
\end{description}
\section{Publications}

Our work in the Instance Selection field has been validated and published in the main Information Systems 
conferences and journals in the last four years only, including 
\textbf{two} published papers in the \textit{Information Processing and Management} \textbf{(IP\&M)} \textit{(h5-index: 96, Impact Factor: 7.4, A1)}, a worldwide leading journal in \textit{Information Retrieval}~\cite{cunha21,cunha20}.
There is also a publication in \textit{ACM Computing Surveys} \textbf{(CSUR)}\textit{(h5-index: 103, IF: 23.8, A1)}~\cite{cunha23}, a 
leading journal in \textit{Computer Science Theory and Methods}, covering comprehensive literature review of both traditional and SOTA methods in the IS field as well as an extensive experimental comparison between these methods; 
the \textit{International Conference \textbf{SIGIR}}~\cite{cunhasigir23}, 
covering our first proposal of a novel redundancy-oriented IS framework
aimed at large datasets with a particular focus on transformers. 
Finally, this dissertation also resulted in an \textit{ACM Transactions on Information Systems} (\textbf{TOIS})~\cite{cunha2024tois} \textit{(h5-index: 48, IF: 5.6, A1)}, a worldwide leading journal in \textbf{Information Systems}, which includes the proposal of an extended noise-oriented and redundancy-aware IS framework for ATC. \looseness=-1

\vspace{-0.2cm}
\begin{enumerate}[leftmargin=*]
\footnotesize
    \item \textbf{Cunha, Washington}, et al. ``A Noise-Oriented and Redundancy-Aware Instance Selection Framework.'' \textbf{ACM Transactions on Information Systems} (ACM TOIS) (2024)-- h5-index: 48.0 
    \item \textbf{Cunha, Washington}, et al. ``An effective, efficient, and scalable confidence-based instance selection framework for transformer-based text classification.'' \textbf{ACM SIGIR} 2023 -- h5-index: 103.0.
    \item \textbf{Cunha, Washington}, et al. ``A Comparative Survey of Instance Selection Methods applied to Non-Neural and Transformer-Based Text Classification.'' \textbf{ACM Computing Surveys} (2023) -- h5-idx: 157.0
    \item \textbf{Cunha, Washington}, et al. ``On the cost-effectiveness of neural and non-neural approaches and representations for text classification:A comprehensive comparative study.''\textbf{IP\&M} (2021) - h5-index: 114.0 
    \item \textbf{Cunha, Washington}, et al. ``Extended pre-processing pipeline for text classification: On the role of meta-feature representations, sparsification and selective sampling.'' \textbf{IP\&M} (2020) -- h5-index: 114.0 
\end{enumerate}

\vspace{-0.2cm}
\noindent Our dissertation contributed \textbf{directly} to other 
journals during the doctorate period:

\vspace{-0.2cm}
\begin{enumerate}[leftmargin=*]
\footnotesize
    \item Gonçalves, M., \textbf{Cunha, W} et al. (2024) ``Mais com menos - processamento de linguagem natural inteligente e sustentável baseado em engenharia de dados e inteligência artificial avançada.'' \textbf{IV Seminário de Grandes Desafios da Computação no Brasil 2025-2035} - Sociedade Brasileira de Computação.
    \item França, C., \textbf{Cunha, W} et al. (2024). On representation learning-based methods for effective, efficient, and scalable code retrieval. \textbf{Neurocomputing}. -- h5-index: 136.0
    \item Andrade, C., \textbf{Cunha, W} (2023) On the class separability of contextual embeddings representations – or “the classifier does not matter when the (text) representation is so good!”. \textbf{IP\&M}. -- h5-index: 114.0
    \item Zanotto, B. S., \textbf{Cunha, W} et al. (2021). Pcv50 automatic classification of electronic health records for a value-based program through machine learning. \textbf{Value in Health}. -- h5-index: 57.0
    \item Zanotto, B. S., \textbf{Cunha, W} et al. (2021). Stroke outcome measurements from electronic medical records: Cross-sectional study on the effectiveness of neural and nonneural classifiers. \textbf{JMIR Med} -- h5-idx: 52.0
    \item Viegas, F. et al., \textbf{Cunha, W} (2024). Exploiting contextual embeddings in hierarchical topic modeling and investigating the limits of the current evaluation metrics.\textbf{Computational Linguistics}. -- h5-idx:38
    \item Felix, L. et al., \textbf{Cunha, W} (2024). Why are you traveling? Inferring trip  profiles from online reviews and domain-knowledge. \textbf{Online Social Networks and Media}. -- h5-index: 28.0
    \item Viegas, F., \textbf{Cunha, W} et al. (2024). Pipelining semantic expansion and noise filtering for sentiment analysis of short documents -- clusent method. \textbf{Journal on Interactive Systems}. -- h5-index: 9.0
    \item Cunha, W et al. (2021) Extended Pre-Processing Pipeline For Text Classification: On the Role of Meta-Features, Sparsification and Selective Sampling. \textbf{CTDBD - SBBD'21} -- h5-index: 7.0
        \item Gomes, C., \textbf{Cunha, W} et al. (2019). CluWords: Explorando Clusters Semânticos entre Palavras para Aprimorar Modelagem de Tópicos. \textbf{CTIC} - Iniciação Científica em Computação. -- h5-index: 1.0

\end{enumerate}

\vspace{-0.2cm}
\noindent Our dissertation contributed \textbf{directly} to other 
conference papers during the Ph.D. period:

\vspace{-0.2cm}
\begin{enumerate}[leftmargin=*]
\footnotesize
\item Viegas, F., \textbf{Cunha, W} et al. (2020). Cluhtm - semantic hierarchical topic modeling based on cluwords. 	Meeting of the Association for Computational Linguistics (ACL) \textbf{ACL'20}. -- h5-index: 215.0
\item Mendes, L., \textbf{Cunha, W} et al. (2020). ``Keep it Simple, Lazy''--- Metalazy: A new metastrategy for lazy text classification. \textbf{CIKM'20}. -- h5-index: 91.0
    \item Viegas, F., \textbf{Cunha, W} et al. (2019). Cluwords: Exploiting semantic word clustering for enhanced topic modeling. \textbf{WSDM'19}. -- h5-index: 77.0
\item Andrade et al., \textbf{Cunha, W} (2024) Explaining the hardest errors of contextual embedding-based classifiers. Conference on Computational Natural Language Learning (\textbf{CoNLL'24}). -- h5-index: 39.0
\item Pasin, A., \textbf{Cunha, W} et al. (2024). A quantum annealing instance selection approach for efficient and effective transformer fine-tuning. ACM SIGIR \textbf{ICTIR}. -- h5-index: 24.0
    \item Fonseca, G., \textbf{Cunha, W} et al. (2024). Estratégias de Undersampling para Redução de Viés em Classificação de Texto Baseada em Transformers. \textbf{WebMedia'24}. -- h5-index: 13.0
    \item Vasconcelos, N., \textbf{Cunha, W} et al. (2024). Integrando Avaliações Textuais de Usuários em Recomendação baseada em Aprendizado por Reforço. \textbf{WebMedia'24}. -- h5-index: 13.0
    \item Viegas, F., \textbf{Cunha, W} et al. (2023). Clusent–combining semantic expansion and de-noising for dataset-oriented sentiment analysis of short texts.  \textbf{WebMedia'23}. -- h5-index: 13.0
    \item Cecilio, P., \textbf{Cunha, W} et al. (2023). Um framework para extração automática de informações em patentes farmacêuticas. \textbf{WebMedia'23}. -- h5-index: 13.0
    \item Júnior, A., \textbf{Cunha, W} et al. (2022) Evaluating topic modeling pre-processing pipelines for portuguese texts. \textbf{WebMedia'22} -- h5-index: 13.0
    \item Pasin, A., \textbf{Cunha, W} et al. (2024). A quantum annealing-based instance selection approach for transformer fine-tuning. \textbf{Italian Information Retrieval Workshop}. -- h5-index: 7.0
    \item Santos, W., \textbf{Cunha, W} et al. (2023). Uma metodologia para tratamento do viés da maioria em modelos de stacking via identificação de documentos difíceis. \textbf{SBBD'23}. -- h5-index: 7.0
    \item Macul, V., \textbf{Cunha, W} et al. (2024). Inteligência Artificial Generativa para Personalização do Cuidado de Saúde Integral: Processo de caracterização dos dados. \textbf{II Simpósio CI-IA Saúde da UFMG}. 

\end{enumerate}
\vspace{-0.3cm}
\section{Systematic Literature Review of Instance Selection Methods}~\label{sec:review}
\vspace{-0.6cm}

\noindent In this section, we present a critical analysis (\textit{a.k.a.,}  {\em rapid (systematic-based) literature review} of the most traditional and/or recent (state-of-the-art) proposals in the Instance Selection (IS) area. This review aims to comprehensively assess the most relevant works related to IS strategies applied in different scenarios. In particular, we focus on experimentally oriented studies, that is, studies that have strong experimental and empirical components to support their findings.\looseness=-1

To achieve our objective, we collected a set of 100 publications\footnote{The list of articles can be found at \url{https://shorturl.at/zCLW7}\label{fn:articlelist}} that included the most cited articles related to IS. We assume that highly cited articles are potentially influential as they have received much attention. From those, we selected the most popular methods to include in our experimental assessment of IS methods applied to  ATC. We also selected a set of recently proposed methods (considered state-of-the-art) to complete our experimental comparison. At the end of this literature review process, which is further detailed next, we end up with  a mix of the traditional and state-of-the-art set of methods, comprising 13 IS strategies to be evaluated in the next section in the ATC scenario.\looseness=-1

\begin{figure}[!h]
    \centering
    \includegraphics[width=0.75\textwidth]{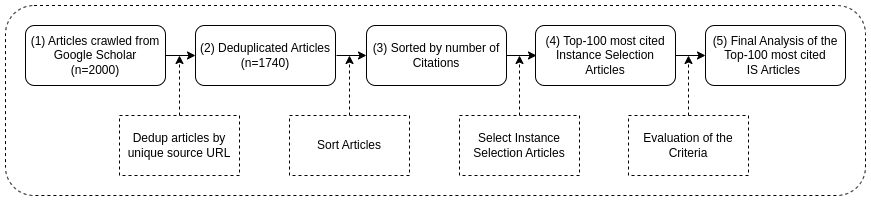}
    \caption{Workflow of study selection and analysis of the (rapid) systematic \\literature review of instance selection methods.}
    \label{fig:review}
\end{figure}

A simplified version of the rapid review procedure is shown in Figure~\ref{fig:review}. First, we collected articles returned by a set of four queries submitted to Google Scholar. Second, we used each article's unique URL to remove duplicates (deduplication phase). This procedure resulted in 1,740 unique articles. Third, we ranked the remaining articles by the number of citations. Fourth, we classified the articles, in ranked order, according to the desired criteria of (i) being related to IS and (ii) conducting experimental comparisons among methods. Fifth, we filtered out articles that did not match the criteria, up to the point that we achieved 100 relevant articles. Up to this point, we had inspection-ed 702 articles in ranked order. From these articles, we selected the most popular and most recent methods to be compared in our experimental assessment. 

The results of our searches and analyses reinforced our perception that IS methods are almost exclusively applied to tabular structured data –- and their application in NLP tasks is rare, which is odd since this is one of the areas that could benefit most from this type of method. In sum, from the 100 considered IS papers, 92 of them considered just tabular data. We propose to investigate the use of IS methods along with ATC models, which have been highly popular in applications as diverse as the detection of fake news and hate speech, sentiment analysis, revision of product characteristics, inferring opinions, and assessing the satisfaction of products and services, among many others.\looseness=-1

We also propose a new and innovative taxonomy of IS strategies by extending a 10-year-old taxonomy proposed in ~\cite{taxonomy}. This previous IS 
taxonomy was proposed in a different context, considering different types of datasets (small tabular ones)  and learning methods. For instance, deep learning neural network methods were not considered in the envisioned scenarios by the time that taxonomy was proposed.

\begin{figure}[!h]
    \centering
    \includegraphics[width=0.75\textwidth]{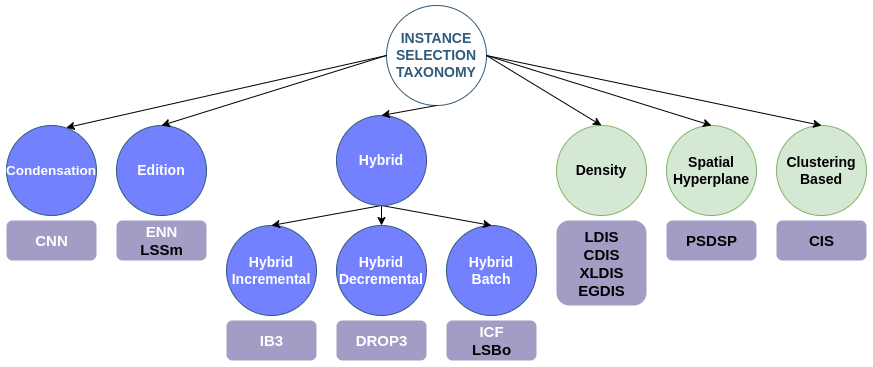}
    \caption{New Extended Instance Selection Taxonomy Proposal}
    \label{fig:taxonomy}
\end{figure}

As mentioned, the field has significantly evolved since the proposal of the original taxonomy. These advances, which we have carefully documented in our new extended taxonomy, are reflected in our new extended taxonomy with three new categories -- represented in green in Figure~\ref{fig:taxonomy} -- and eight new, recently proposed methods -- written with black color letters in Figure~\ref{fig:taxonomy}.
The new categories refer to approaches based on density, spatial hyperplanes, and clustering-based approaches proposed since 2015.\looseness=-1

Last, in our Ph.D. Dissertation, we also provided detailed information about the thirteen IS approaches evidenced in our critical analysis (not presented here due to space constraints) that we will consider in our experimentation evaluation for the ATC context. Therefore, in the next section, we provide a complete experimental evaluation comparing the IS methods detailed above, combining them with SOTA ATC strategies. Our objective with this set of experiments is to answer the first posed research question (RQ1).\looseness=-1
\section{A Comparative Survey of Instance Selection Methods applied to NonNeural and Transformer-Based Text Classification}

In this section, we propose to assess the trade-off among reduction, efficiency, and effectiveness of these 13 most representative traditional IS methods (Section~\ref{sec:review}) applied to the ATC task using large and varied datasets. It is essential to notice that the selected IS methods presented in the previous section have been tested only with small structured tabular datasets (such as those from the UCI repository). Regarding ATC methods, we considered the current SOTA in the ATC field: the transformer-based architectures, such as BERT, XLNet, RoBERTa, and others. These methods have a high cost regarding computational resources, mainly when dealing with large labeled training data. Therefore, they constitute an ideal scenario for the application of IS techniques. Last, but not least important, we should emphasize that our work, as far as we know, is the first to apply IS as a preprocessing step before using transformer-based architectures in the ATC context. This contribution is realized using an experimental study whose rigor and magnitude (7 transformer methods and 13 IS approaches) have not yet been reported in the literature on IS.\looseness=-1

Next, we present the comparative Scenario and experimental setup proposed to assess the trade-off among reduction, efficiency, and effectiveness of these 13 most representative traditional IS methods applied to the ATC.\looseness=-1

\begin{table}[!h]
    \centering
    \scriptsize
    \begin{tabular}{llrrrrr}
        Task & Dataset       & Size      & Dim. & \# Classes  & Density  & Skewness    \\
        \midrule
\parbox[t]{1mm}{\multirow{11}{*}{\rotatebox[origin=c]{90}{Topic}}}  & DBLP & 38,128 & 28,131 & 10 & 141 & Imbalanced \\
 & Books & 33,594 & 46,382 & 8 & 269 & Imbalanced \\
 & ACM & 24,897 & 48,867 & 11 &  65 & Imbalanced   \\
 & 20NG & 18,846 & 97,401 & 20 &  96 & Balanced    \\
 & OHSUMED & 18,302 & 31,951 & 23 & 154 & Imbalanced   \\
 & Reuters90 & 13,327 & 27,302 & 90 & 171 & Extremely   Imbalanced \\
 & WOS-11967 & 11,967 & 25,567 & 33 & 195 & Balanced \\
 & WebKB & 8,199 & 23,047 & 7 & 209 & Imbalanced   \\
& Twitter & 6,997 & 8,135 & 6 & 28 & Imbalanced \\ 
 & TREC & 5,952 & 3,032 & 6 & 10 & Imbalanced \\
 & WOS-5736 & 5,736 & 18,031 & 11 & 201 & Balanced \\ \hline
\parbox[t]{1mm}{\multirow{8}{*}{\rotatebox[origin=c]{90}{Sentiment}}}  & SST1 & 11,855 & 9,015 & 5 & 19 & Balanced         \\
 & pang\_movie & 10,662 & 17,290 & 2 & 21 & Balanced        \\
 & Movie Review & 10,662 & 9,070 & 2 & 21 & Balanced   \\
 & vader\_movie & 10,568 & 16,827 & 2 & 19 & Balanced        \\
 & MPQA & 10,606 & 2,643 & 2  & 3 & Imbalanced         \\
 & Subj & 10,000 & 10,151 & 2 & 24 & Balanced         \\
 & SST2 & 9,613 & 7,866 & 2 & 19 & Balanced         \\
 & yelp\_reviews & 5,000 & 23,631 & 2 & 132 & Balanced        \\
\hline
\parbox[t]{1mm}{\multirow{3}{*}{\rotatebox[origin=c]{90}{Large}}} & AGNews & 127,600 & 39,837 & 4 & 37 & Balanced \\
& Yelp\_2013 & 335,018 & 62,964 & 6 & 152 & Imbalanced \\
& MEDLINE &	 860,424 & 125,981 & 7 & 77 & Extremely Imbalanced \\

        \bottomrule
    \end{tabular}
    \caption{Datasets Statistics, including: size, dimensionality, classes, document density, and class distribution.}
    \label{tbl:datasets_stats}
    \vspace{-0.8cm}
\end{table}

\paragraph*{Datasets, Data Representation, and Preprocessing}
We adopted the 22 real-world datasets collected from various sources in two broad ATC tasks~\cite{li2022survey}: i) \textit{topic classification}; and ii) \textit{sentiment analysis}. The datasets cover a range of domains, diversity in size, dimensionality, classes, document density, and class distribution.\looseness=-1

The TFIDF representation is input to all IS methods.
Other text representations could have been adopted as input to IS methods, such as \emph{static embeddings} (e.g., FastText~\cite{joulin2016fasttext}) or \emph{contextualized embeddings} built by transformer architectures (whether by forwarding documents through fine-tuned model or a zero-shot approach). However, as previously demonstrated in~\cite{andrade2024explaining}: (i) static embeddings can slow down classification methods significantly; and (ii) using contextualized embeddings directly as IS input is inefficient and ineffective, probably because it sacrifices sparsity. In this context, given the goal of finding cost-effective solutions, the TF-IDF representation is still the best alternative as input for the instance selection methods.\looseness=-1

\begin{figure}[h]
\centering
\includegraphics[width=0.75\textwidth]{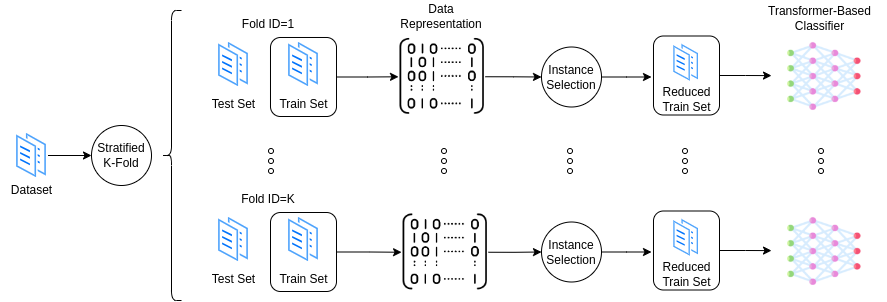}
\caption{Data Representation and Preprocessing Procedure}
\label{fig:IS_data_rep_and_pre}
\vspace{-1.0cm}
\end{figure}

We removed stopwords and kept features appearing in at least two documents. We normalized the TF-IDF product result using the L2-norm. In practice, as illustrated in \textbf{Figure~\ref{fig:IS_data_rep_and_pre}},  we first split the dataset employing the Stratified K-Fold cross-validation methodology -- the smaller datasets were executed using k=10-fold partition, while for the larger ones, we adopted 5 folds due to the cost of the procedure --, then we construct the TFIDF matrix representation of the documents for the IS stage, and then, we use the corresponding raw document chosen as input for the Transformers classifiers.\looseness=-1

\vspace{-0.3cm}

\paragraph{Text Classification Methods} 

As mentioned, our goal is to study and compare our proposed method against the SOTA IS techniques in the context of \textbf{Transformers} architectures  -- notably the SOTA in classification in several domains\footnote{LLMs such as GPT and LLama are built on top of Transformer architectures.}~\cite{Gasparetto2022ASO,de2024strategy}. 
We compare the effectiveness among the latest version of the following Transformers --  \textbf{RoBERTa}, \textbf{BERT}, \textbf{DistilBERT}, \textbf{BART}, \textbf{AlBERT}, and \textbf{XLNet}) -- applied to all tested datasets. \looseness=-1

Given the large number of hyperparameters to be tuned, performing a grid search with cross-validation is not feasible for all of them. As a result, to determine the optimum hyperparameter, we applied the methodology from \cite{cunha21}. We fixed the initial learning rate as $5 \times 10^{-5}$, the max number of epochs as $20$, and $5$ epochs as patience. Finally, we perform a grid search on max\_len (150 and 256) and batch\_size ($16$ and $32$) since these specified values directly impact efficiency and effectiveness.\looseness=-1

\vspace{-0.5cm}
\paragraph{Instance Selection Methods}

We consider in this chapter a set of 13 IS methods
described in Section~\ref{sec:review}, namely: 
 \textit{Condensed Nearest Neighbor} (\textbf{CNN}); \textit{Edited Nearest Neighbor} (\textbf{ENN}); \textit{Iterative Case Filtering} (\textbf{ICF});
 \textit{Instance Based 3} (\textbf{IB3}); \textit{Decremental Reduction Optimization Procedure} (\textbf{Drop3}); \textit{Local Set-based Smoother} (\textbf{LSSm}); \textit{Local Set Border Selector} (\textbf{LSBo}); \textit{Local Density-based Instance Selection} (\textbf{LDIS}); \textit{Central Density-based Instance Selection} (\textbf{CDIS}); \textit{eXtended Local Density-based Instance Selection} (\textbf{XLDIS}); \textit{Prototype Selection based on Dense Spatial Partitions} (\textbf{PSDSP});
 \textit{Enhanced Global Density-based Instance Selection} (\textbf{EGDIS}); and \textit{Curious Instance Selection} (\textbf{CIS}). 
 All parameters for the IS methods were defined with grid-search, using cross-validation in the training set during an initial empirical experiments round.

\paragraph*{Metrics and Experimental Protocol}
All experiments were executed on an Intel Core i7-5820K with 6-Core and 12-Threads, running at 3.30GHz, 64Gb RAM, and a GeForce GTX TITAN X (12GB) and Ubuntu 19.04. Due to skewness in the datasets, we evaluated the classification effectiveness using Macro Averaged F1 (MacroF1)\cite{Sokolova}. We employed the paired t-test with a 95\% confidence level to compare the average outcomes from our cross-validation experiments. Finally, we applied the Bonferroni correction~\cite{bonferroni} to account for multiple tests. \looseness=-1
We consider reduction mean by defined as 
$
    \overline{R} = \frac{\sum_{i=0}^{k} \frac{|T_i|-|S_i|}{|T_i|}}{k}
$, 
where $T$ is the original training set, and $S$ is the solution set containing the selected instances by the IS method being evaluated. \looseness=-1
Last, in order to analyze the cost-effectiveness tradeoff, we also evaluate each method's cost in terms of the total time required to build the model. The Speedup is calculated as $
    S = \frac{T_{wo}}{T_w}
$,
where $T_w$ is the total time spent on model construction using the IS approach, 
and $T_{wo}$ is the total time spent on execution without the IS phase. \looseness=-1

\noindent \textbf{Experimental Results:}
Our findings reveal a significant potential of IS. In some cases, these methods can reduce the training set by up to 90\% while maintaining their effectiveness. The IS approaches we studied achieved average reductions between 15.6\% (LSSm) to 91.1\% (XLDIS). However, despite the motivation for noise removal, selection methods were not able to improve the effectiveness of the text classification models.
This underscores the need for further studies to investigate this issue, which could have profound implications for the field of data science and machine learning.\looseness=-1

We demonstrated that three traditional IS methods \textbf{(LSSm, CNN, LSBo)} were able to reduce the total text classification models' construction time while keeping the effectiveness in 12 (out of 19) considered datasets -- with speedups between 1.04x (CNN - Books) and 5.69x (LSBo - Reuters90). In the other datasets, we observed that the introduction of IS approaches caused an overhead\footnote{We fully explore this time overhead behaviour in our Dissertation (Chapter 3.3.3)} in terms of the total time to generate the model (running the IS methods + model construction), making the whole process more costly from a computational cost perspective. Overall, considering the three tripod constraints altogether and all datasets, the best traditional IS method was \textbf{CNN}.

Answering RQ1, our evaluation of the tripod constraints (reduction - efficiency - effectiveness), we showed that, in some datasets, specific selection methods can reduce the training set without loss of effectiveness and with efficiency improvements. \textbf{However, our experiments revealed that no IS method can respect all restrictions in all cases.} Also, in some situations, the application of the IS methods can incur additional overheads in the time to construct the model. Our results present a partial answer to the posed question regarding the need for large training sets for performing fine-tuning. In some cases, it is possible to use IS methods to reduce the training set without loss of effectiveness and efficiency gains -- meaning that we do not always need a lot of data. Therefore, these findings neither totally support nor completely refute our posed hypothesis that traditional IS methods are capable of simultaneously respecting all posed restrictions.\looseness=-1

\begin{table*}[h]
    \centering
    \small
    \resizebox{0.9\textwidth}{!}{
    \begin{tabular}{l|l}
        \toprule
        \multirow{3}{*}{\rotatebox[origin=c]{90}{\textbf{CNN}}} & \textbf{(P)}: Best overall method, achieving the best trade-off on the tripod Reduction-Effectiveness-Efficiency\\
 & \textbf{(C)}: Limited effectiveness  when applied to the largest datasets \\
 & \textbf{(R)}:  Medium-to-small topic-related data or sentiment analysis tasks\\\hline

        \multirow{3}{*}{\rotatebox[origin=c]{90}{\textbf{LSSm}}} & \textbf{(P)}: Achieved the best overall Effectiveness\\
 & \textbf{(C)}: Lowest overall Efficiency when considering the three best IS approaches\\
 & \textbf{(R)}: General tasks that cannot deal with effectiveness losses\\\hline
\multirow{3}{*}{\rotatebox[origin=c]{90}{\textbf{LSBo}}} & \textbf{(P)}: Considering the three best IS methods in terms of effectiveness, achieved the best reduction rate\\
 & \textbf{(C)}: Limited results considering effectiveness when applied to the topic-related tasks\\
 & \textbf{(R)}: Sentiment analysis tasks that need performance and scalability \\\hline
\multirow{3}{*}{\rotatebox[origin=c]{90}{\textbf{EGDIS}}} & \textbf{(P)}: Considering the IS methods in this list, achieved the best reduction rate and speed up trade-off\\
 & \textbf{(C)}: Limited results considering effectiveness when applied to the medium-to-large topic-related task\\
 & \textbf{(R)}: Tasks that need performance and scalability and may afford some (up to 5\%)  effectiveness losses\\\hline
\multirow{3}{*}{\rotatebox[origin=c]{90}{\textbf{CIS}}} & \textbf{(P)}: Good effectiveness results in sentiment analysis tasks\\
 & \textbf{(C)}: Lowest overall Efficiency\\
 & \textbf{(R)}: Not recommended for large NLP tasks due to high associated computational cost\\\hline
\multirow{3}{*}{\rotatebox[origin=c]{90}{\textbf{IB3}}} & \textbf{(P)}: $4^{th}$ best IS method considering effectiveness \\
 & \textbf{(C)}: Limited results considering effectiveness when applied to large topic-related tasks\\
 & \textbf{(R)}: Medium-to-small topic-related data or sentiment tasks that can deal with a limited effectiveness loss\\

        \bottomrule
    \end{tabular}
    }
    \caption{{Instance Selection Pros~\textbf{(P)}, Cons~\textbf{(C)} and Recommendations~\textbf{(R)}.}}
    \label{tbl:final_summ}
\end{table*}

In Table \ref{tbl:final_summ}, we provide pros~\textbf{(P)}, cons\textbf{(C)} and recommendations~\textbf{(R)} for the six best IS methods. The remaining seven methods were ineffective in the ATC context, producing statistically worse results in 116 out of 133 (7 methods x 19 datasets). Particularly our focus is on NLP tasks, especially ATC ones.\looseness=-1

Our results motivate further investigations on exploiting IS methods in the ATC context, especially regarding new transformers. Our study concerning RQ1 also opens space for designing new, more efficient, effective, and scalable IS methods for the current ATC and the big data scenarios in general. To help close this gap, in the next chapter, we introduce the E2SC framework, which is a new two-step framework that satisfies all the constraints of the tripod and can be used in real-world situations, even with datasets containing thousands of instances, with a special focus on transformer-based architectures. \looseness=-1

\noindent \blue{\textbf{Final Remark:} Both sections 4 and 5 played a crucial role in producing an impactful publication 
in the \textit{ACM Computing Surveys} \textbf{(CSUR)}~\cite{cunha23}. }
\section{An Effective, Efficient, and Scalable Confidence-Based Instance Selection Framework for Transformer-Based Text Classification}

In the previous section, we found that no one method was able to meet all restrictions in all cases. Additionally, in some scenarios, using these methods increased the time it takes to construct the model, which corresponds to a gap in the literature on methods capable of respecting the posed restriction simultaneously.\looseness=-1

\begin{figure}[h]
\centering
\includegraphics[width=0.6\textwidth]{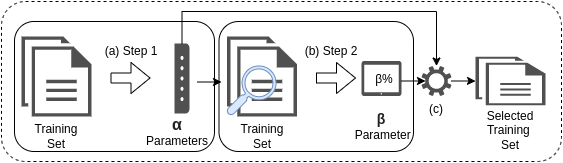}
\caption{The proposed E2SC Framework.}
\label{e2scISMethod}
\end{figure}

To help close this gap, the main contribution of this section is the proposal of \textbf{E2SC} -- \underline{\textbf{E}}ffective, \underline{\textbf{E}}fficient, and \underline{\textbf{S}}calable \underline{\textbf{C}}onfidence-based \underline{\textbf{I}}nstance \underline{\textbf{S}}election -- a novel two-step framework\footnote{To guarantee the reproducibility of our solutions, all the code, the documentation of how to run it and datasets are available on: \url{https://github.com/waashk/}} aimed at large datasets with a special focus on transformer-based architectures, our first redundancy-oriented IS solution. E2SC is a technique that satisfies the tripod´s constraints and is applicable in real-world scenarios, including datasets with thousands of instances. E2SC´s overall structure can be seen in Figure~\ref{e2scISMethod}.

E2SC's \textbf{first} step -- Figure~\ref{e2scISMethod}(a) -- aims to assign a probability to each instance being removed from the training set ($\alpha$ parameters). We adopt an exact KNN model solution\footnote{We depart from the premise that the exact KNN model solution is a reasonable proxy for redundancy. We test and confirm this premise in our Dissertation.} to estimate the probability of removing instances, as it is considered a \textbf{calibrated}\footnote{A calibrated classifier is one whose probability class predictions correlate well with its accuracy, e.g., for those instances predicted with 80\% of confidence the classifier is correct in roughly 80\% of the cases.}\cite{rajaraman2022deep} and computationally cheap (fast) model~\cite{cardoso2017ranked}. Our \textbf{first hypothesis (H1)} is that high confidence (if the model is calibrated to the correct class, known in training) positively correlates with redundancy for the sake of building a classification model. Accordingly,  we keep the hard-to-classify instances  (probably located in the decision border regions), weighted by confidence, for the next step, in which we partially remove only the easy ones.\looseness=-1

As the \textbf{second} step of our method -- Figure~\ref{e2scISMethod}(b) -- we propose to estimate a near-optimal reduction rate ($\beta$ parameter) that does not degrade the deep model's effectiveness by employing a validation set and a weak but fast classifier. Our \textbf{second hypothesis (H2)} is that we can estimate the effectiveness behavior of a robust model (deep learning) through the analysis and variation of selection rates in a weaker model. Again, we explore KNN for this. More specifically, we introduce an iterative method that statistically compares, using the validation set, the KNN model's effectiveness without any data reduction against the model with iterative data reduction rates. In this way, we can estimate a reduction rate that does not affect the KNN model's effectiveness. Last, considering the output of these two steps together (Figure~\ref{e2scISMethod}(c)), $\beta\%$ instances are randomly sampled, weighted by the $\alpha$ distribution, to be removed from the training set.\looseness=-1

\vspace{0.6cm}
\noindent \textbf{Experimental Results:}

We compare E2SC proposal with \textbf{six} robust state-of-the-art instance selection baseline methods considering as input of the best of \textbf{seven} deep learning text classification methods in a large benchmark with \textbf{19} datasets. Our experimental evaluation shows that \textbf{E2SC} managed to significantly reduce the training sets (by \textbf{27\%} on average; varying between 10\% and 60\% of reduction) while maintaining the same levels of effectiveness in \textbf{18} (out of 19) considered datasets. Also, we found that \textbf{E2SC} was able to reduce the total ATC models' construction time while keeping the effectiveness in all (19) considered datasets -- with speedups of \textbf{1.25} on average, varying between 1.02x (Books) and 2.04x (yelp\_reviews). Overall, considering the three tripod constraints altogether and all datasets, the best IS method so far was our first proposed framework. Finally, to demonstrate the flexibility of our framework to cope with large datasets, we propose two modifications. Our enhanced solution managed to increase the reduction rate of the training sets (to \textbf{29\%} on average) while maintaining the same levels of effectiveness in \textbf{all} datasets, with speedups of \textbf{1.37} on average. The framework scaled to large datasets, reducing them by up to 40\% while statistically maintaining the same effectiveness with speedups of \textbf{1.70x}.\looseness=-1

Despite being innovative and achieving significant results in terms of effectiveness, efficiency, and reduction, the E2SC framework focused only on \textbf{redundancy}, leaving some other aspects that may help to reduce training further untouched.
One such aspect is \textbf{noise}, here defined as instances incorrectly labeled by humans in the dataset~\cite{whyfail} as well as (possible) outliers that do not contribute (or even get in the way) to model learning (tuning). Indeed, according to~\cite{whyfail}, users, whether regular individuals or experts, make a reasonable amount of mistakes while labeling complex instances -- between 56\% and 64\% of the time. Other few noisy instances (a.k.a., outliers) may be correctly labeled, but they differ so significantly from other instances from the same class that they are either useless for the sake of learning (tuning) or may even be detrimental to the process.\looseness=-1

Noisy instances can potentially constitute a significant portion of available data in these contexts. Noisy (training) instances may not only degrade the model's effectiveness by incorporating misleading patterns in the model but may also be detrimental to performance as they need to be processed to extract and incorporate these patterns into the model. If the amount of noise is significant, there will certainly be negative impacts on effectiveness and efficiency. However, in a simulated scenario designed to evaluate the capability of the IS baseline methods and our previous solution to remove noise, none of the IS solutions satisfactorily performed the task.\looseness=-1 

\noindent \textbf{Final Remark:} This section was summarized in an conference paper in the \textit{ACM SIGIR Conference on Research and Development in Information Retrieval}~\cite{cunhasigir23}. 
\section{An Extended Noise-Oriented and Redundancy-Aware Instance Selection Framework for Transformer-Based Automatic Text Classification}

The main contribution of this section is the proposal of an \textit{extended bi-objective instance selection (\textbf{biO-IS})} framework built upon our first one aimed at removing both redundant and noisy instances simultaneously.

\begin{figure}[!h]
\centering
\includegraphics[width=0.8\textwidth]{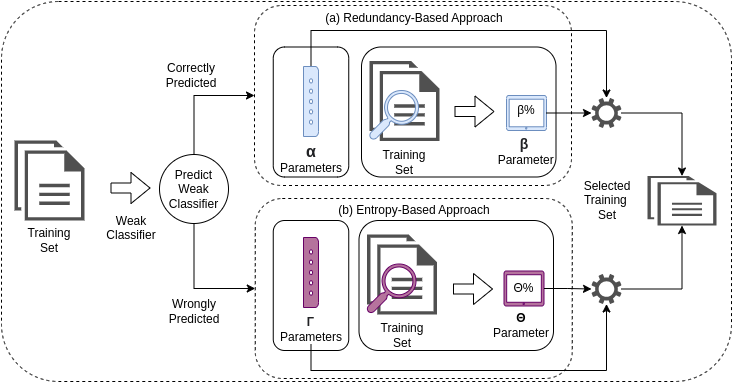}
\caption{Bi-objective Instance Selection Framework}
\label{fig:tois:solution}
\end{figure}

The main contribution of this section is the proposal of an \textit{extended bi-objective instance selection (\textbf{biO-IS})} framework built upon our first one aimed at removing both redundant and noisy instances simultaneously. As depicted in Figure~\ref{fig:tois:solution}, our extended framework encompasses three main components: a weak classifier, a redundancy-based approach, and an entropy-based approach. In Figure~\ref{fig:tois:solution} (in blue), we depart from our original solution proposed in the previous section considering the Logistic Regression as the calibrated weak classifier instead of KNN, as in our original work. An in-depth comparative analysis of several possibilities of the weak classifiers (see more in the Dissertation), including Decision Tree (DT), Logistic Regression (LR), XGBoost, LightGBM (LGBM), and Linear SVM, demonstrated LR as the best option in terms of a trade-off effectiveness-calibration-cost.\looseness=-1 

To address the second objective of noise removal (the lower part of Figure ~\ref{fig:tois:solution} – in purple), we propose a new step based on the entropy and a novel iterative process to estimate near-optimum reduction rates. Considering the instances wrongly predicted by the weak classifier, the main objective is to assign a probability to each of them being removed from the training set based on the probability of the instance being noise. For this, we propose using the entropy function as a proxy to determine the reduction behavior caused by these instances for the sake of training an ATC model. The intuition behind this new step is that when the prediction provided by the calibrated weak classifier is incorrect,  the entropy of the posterior probabilities negatively correlates with the classifier's confidence. This means that low entropy occurs when the classifier assigns an instance with absolute certainty to a wrong class, while high entropy occurs when the classifier is uncertain among several classes. Therefore, we consider the chance of a noisy instance being removed by the inverse of the entropy of the prediction, {so when an incorrect prediction is accompanied by low entropy, it is more likely to be removed, and, otherwise, it is more likely to be kept.} Accordingly, the proposed biO-IS framework provides a comprehensive solution to address redundancy and noise removal simultaneously.\looseness=-1 

\noindent \textbf{Experimental Results:}

We compare \textbf{biO-IS} with \textbf{seven} robust SOTA IS baseline methods, including our first proposal, E2SC, in the text classification domain considering the same benchmark covering \textbf{22} datasets. Our experimental evaluation reveals that, in a simulated scenario designed to evaluate the capability of the IS baseline methods and our previous solution to remove noise, none of the IS solutions were capable of satisfactorily performing the task. On the other hand, \textbf{biO-IS} managed to remove up to 66.6\% of the manually inserted noise. Moreover, \textbf{biO-IS} managed to significantly reduce the training sets (by \textbf{40.1\%} on average; varying between 29\% and 60\% of reduction) while maintaining the same levels of effectiveness in \textbf{all} of the considered datasets. Also, \textbf{biO-IS} managed to consistently provide speed-ups of \textbf{1.67x} on average (maximum of \textbf{2.46x}). No baseline, not even our previous SOTA solution, was capable of achieving results with this level of quality, considering all tripod criteria. Indeed, the only other method capable of maintaining the effectiveness on all datasets was E2SC; \textbf{biO-IS} improves over E2SC by 41\% regarding reduction rate and from 1.42 to 1.67 (on average) regarding speedup, achieving the SOTA in the Instance Selection field.\looseness=-1

\noindent \blue{\textbf{Final Remark:} This section was summarized in a  journal paper in the \textit{ACM Transactions on Information Systems}~\cite{cunha2024tois}, a 
leading journal in \textbf{Information Systems}. }
\section{Conclusion and Future Work}

This dissertation surveyed classical and recent IS approaches, revealing significant advances but limited application scope. Most traditional methods target small tabular datasets, with rare applications in NLP despite its potential benefits.\looseness=-1

To address this gap, we conducted a rigorous comparative study of classical and SOTA IS methods applied to ATC. The study evaluated the trade-offs among reduction, effectiveness, and cost, motivated by the rising costs of new ATC solutions due to deep learning embeddings and increasing data volumes. The main findings from over 5,000 experiments include: (i) IS methods rarely improved ATC model performance, with a few exceptions like XLNet. (ii) Among 13 IS methods, LSSm, CNN, and LSBo performed best, achieving significant reductions (46.6\% average) while maintaining effectiveness in 12 datasets. (iii) Fine-tuning transformers proved critical for effectiveness, and IS methods helped reduce training costs in many cases (161 out of 190). Contrary to common beliefs, neural networks often require representative—not large—data to perform well in ATC. Overall, IS techniques effectively reduce training set sizes without compromising effectiveness, especially in fine-tuning transformer models. However, traditional approaches often fall short of meeting all trade-offs, underscoring the need for more efficient, scalable IS techniques in big data scenarios.\looseness=-1

To address these challenges, we proposed E2SC, a redundancy-oriented two-step IS framework designed for large datasets and transformer-based architectures, introducing: (i) Calibrated weak classifiers to estimate data usefulness during transformer training. (ii) Iterative processes and heuristics to determine optimal reduction rates. E2SC reduced training sets by 30\% on average while maintaining effectiveness across 22 datasets, achieving speedups of up to 70\%.\looseness=-1

To overcome E2SC limitations regarding handling noise effectivly, we developed biO-IS, an extended framework that simultaneously removes redundant and noisy instances. Experimental results showed biO-IS reduced training sets by 41\% on average (up to 60\%), removed 66.6\% of manually inserted noise, and maintained effectiveness across datasets. It provided consistent speedups (1.67x average, up to 2.46x), outperforming E2SC in reduction and efficiency.\looseness=-1 

These findings confirm the dissertation hypothesis: small and large language models can be trained with less data without sacrificing effectiveness. This not only enables significant cost and energy savings but also contributes to reducing carbon emissions. Such promising results instill hope for a more sustainable and efficient future in NLP, where advancements in IS techniques can lead to substantial environmental and economic benefits.\looseness=-1 

By proposing and evaluating the proposed enhancements (modifications and extensions) that resulted in the proposed \emph{biO-IS} framework, our work opens several new opportunities. {In future work, we intend to evaluate our proposed framework on other models, such as T5 and DeBERTa, assessing its generalizability.} We also plan to apply IS concepts in other fields beyond ATC, such as searching, ranking, recommendation, and other NLP tasks, such as question answering and supervised topic modeling. So far, we have measured the importance of a whole document for the training (tuning) process, but in the future, we could do this at a more fine-grained level, such as for reviews' aspects~\cite{cunha18} and documents' passages. We also plan to investigate using our framework in an unlabeled deep-learning pre-training stage, e.g., to build a large language model from scratch more efficiently.\looseness=-1 

We also want to investigate the introduction of IS techniques in the context of AutoML solutions as a step in a pipeline of transformations as in~\cite{cunha20}. Furthermore, considering the vast number of dimensions in textual datasets, assessing how \emph{feature selection} methods interact with \emph{instance selection} can be very interesting. Essentially, this involves working with both the rows (documents) and the columns (features) of a document-feature matrix representing textual data. Last but not least, we intend to perform a cost-benefit analysis on LLMs (effectiveness vs. cost) to potentially pursue IS for fine-tuning LLMs, also exploiting new scalable computational paradigms such as quantum computing~\cite{ferrari2024quantum}.\looseness=-1

\vspace{-0.4cm}
\paragraph*{Acknowledgements}
This work was partially supported by CNPq, CAPES, INCT-TILD-IAR, FAPEMIG, AWS, Google, NVIDIA, CIIA-Saúde, and FAPESP. 
\vspace{-0.1cm}

\footnotesize
\bibliographystyle{sbc}
\bibliography{main}

\end{document}